\documentclass[10pt,twocolumn,letterpaper]{article}

\usepackage[pagenumbers]{cvpr} 

\usepackage{graphicx}
\usepackage{amsmath}
\usepackage{amssymb}
\usepackage{booktabs}

\usepackage{stackengine}

\usepackage{blindtext}
\usepackage{amsmath}
\usepackage{multirow}

\usepackage{bm}
\usepackage{bbm} 
\usepackage{comment}
\usepackage{algorithm}
\usepackage{algorithmic}
\usepackage{array} 
\usepackage[dvipsnames]{xcolor}
\usepackage[safe]{tipa} 
\usepackage{color, colortbl}
\usepackage[misc,geometry]{ifsym} 
\usepackage[pagebackref,breaklinks,colorlinks]{hyperref}
\makeatletter

\usepackage[capitalize]{cleveref}
\crefname{section}{Sec.}{Secs.}
\Crefname{section}{Section}{Sections}
\Crefname{table}{Table}{Tables}
\crefname{table}{Tab.}{Tabs.}

\def\cvprPaperID{523} 
\def\confName{CVPR}
\def\confYear{2022}

\begin{document}
	\definecolor{Gray}{gray}{0.95}
	
	\definecolor{LightCyan}{rgb}{0.88,1,1}
	\title{SIGMA: Semantic-complete Graph Matching for Domain Adaptive Object Detection}
	
	\author{Wuyang Li \qquad Xinyu Liu \qquad Yixuan Yuan\thanks{Yixuan Yuan is the corresponding author. \newline This work was supported by Hong Kong Research Grants Council (RGC) General Research Fund 11211221 (CityU 9043152).}\\
		City University of Hong Kong\\
		{\tt\small \href{mailto:wuyangli2-c@my.cityu.edu.hk}{\{wuyangli2,}
			\href{mailto:xliu423-c@my.cityu.edu.hk}{xliu423\}}\href{mailto:wuyangli2-c@my.cityu.edu.hk}{-c@my.cityu.edu.hk}  \href{mailto:yxyuan.ee@cityu.edu.hk}{yxyuan.ee@cityu.edu.hk}}
	}
	\maketitle
	
	
	\begin{abstract}
		
		Domain Adaptive Object Detection (DAOD) leverages a labeled domain to learn an object detector generalizing to a novel domain free of annotations. Recent advances align class-conditional distributions by narrowing down cross-domain prototypes (class centers). Though great success, they ignore the significant within-class variance and the domain-mismatched semantics within the training batch, leading to a sub-optimal adaptation. To overcome these challenges, we propose a novel SemantIc-complete Graph MAtching (SIGMA) framework for DAOD, which completes mismatched semantics and reformulates the adaptation with graph matching. Specifically, we design a Graph-embedded Semantic Completion module (GSC) that completes mismatched semantics through generating hallucination graph nodes in missing categories. Then, we establish cross-image graphs to model class-conditional distributions and learn a graph-guided memory bank for better semantic completion in turn. After representing the source and target data as graphs, we reformulate the adaptation as a graph matching problem, i.e., finding well-matched node pairs across graphs to reduce the domain gap, which is solved with a novel Bipartite Graph Matching adaptor (BGM). In a nutshell, we utilize graph nodes to establish semantic-aware node affinity and leverage graph edges as quadratic constraints in a structure-aware matching loss, achieving fine-grained adaptation with a node-to-node graph matching. Extensive experiments verify that SIGMA outperforms existing works significantly. Our code is available at \url{https://github.com/CityU-AIM-Group/SIGMA}.
		
	\end{abstract}

	
	\section{\label{sec:intro}Introduction}
	
	Well-trained object detectors~\cite{fasterrcnn,fcos,yolov3} have been proven to achieve promising performance with a consistent distribution of training and test data. However, deploying these methods in a novel domain leads to the catastrophic performance degradation due to the domain gap~\cite{DAfasterrcnn}, which significantly limits the generalization and transferability of object detectors. Furthermore, this challenge also restricts the application of object detection in real-world scenarios, such as self-driving under distinctive weather conditions and video analysis containing novel scenes.
	
	\begin{figure}[t]
		\begin{center}
			\includegraphics[width=1.0\linewidth]{ 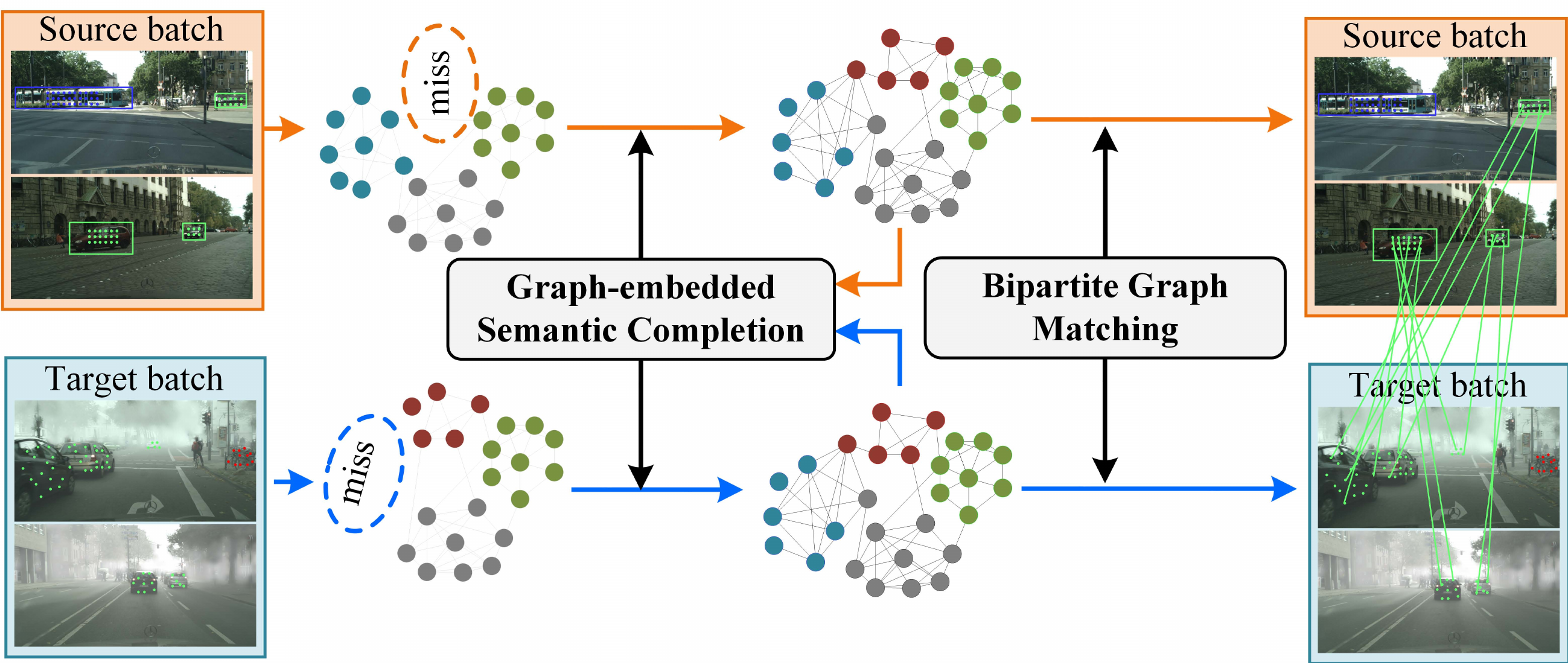}
		\end{center}
		\vspace{-15pt}
		\caption{Illustration of the proposed SemantIc-complete Graph MAtching (SIGMA) framework for DAOD.}
		\vspace{-10pt}
		\label{motivation}
	\end{figure}
	
	To overcome this limitation, Unsupervised Domain Adaptation (UDA) methods have been explored to adapt the unlabeled target domain and the annotated source domain, and one of the main streams of UDA works is to align feature distributions between source and target domains. Early works~\cite{DAfasterrcnn,everypixelmatters,swda} adopt a pixel-to-pixel adaptation in terms of hierarchical features, yielding a global alignment of the whole image with per-pixel adaptation. Some works~\cite{DAfasterrcnn,xu2020crossgraph,RPN} focus on foreground objects and conduct more precise adaptation on those regions of interest. Recently, some works~\cite{xu2020crossgraph,c2f,RPN,KTNet,bipartite_graph} aim to align the cross-domain class-conditional distribution in the implicit feature space and achieve adaptation in a category-to-category manner. These works model category centers with prototypes and minimize the distance of cross-domain prototypes to bridge the domain gap at the category level.
	
	Though satisfactory performance, there are still two challenges in existing category-level adaptation works~\cite{xu2020crossgraph,c2f,RPN,KTNet}. Firstly, these works neglect the significant within-class variance and directly align handcraft category centers, which inevitably bring about a sub-optimal adaptation. Due to the diverse size and appearance of object instances, the within-class variance covers essential information to represent class-conditional distributions, e.g., the scale and shape, which should also be aligned for domain adaptation. Overlooking the within-class variance could lead to lots of non-adapted object instances and the potential overlapping of different class-conditional distributions with false-positive classification errors. Although some works have introduced explicit variance~\cite{var} to relieve the problem of existing center-based measurements, they follow the Gaussian assumption to model feature distributions, which is not optimal in the non-convex deep feature space. These observations motivate us to design a new paradigm to align cross-domain pixel-pairs in the non-euclidean graphical space~\cite{graph_da}, which models and adapts class-conditional distributions without handcraft center-based alignment.
	
	The second challenge lies in the domain-mismatched semantics within the training batch. Some existing works~\cite{xu2020crossgraph,KTNet,c2f} only perform adaptation on the co-occurred categories in two domains, ignoring mismatched categories appearing in a single domain. Neglecting missing categories leads to a non-effective adaptation due to the loss of semantic knowledge. As shown in Figure~\ref{motivation}, the \textcolor{blue}{train} only appears in the source batch, while these \textcolor{red}{bicycles} are available in the target domain, yielding inconsistent semantics across domains. These mismatched semantics bring about the difficulty of explicitly estimating class centers, limiting the adaptation of class-conditional distributions. Furthermore, the missing semantics in the target domain even result in the potential risk towards source-specific direction since the supervised source classification could generate a biased class-conditional distribution~\cite{DSS}. Hence, we are committed to designing a semantic completion strategy through generating novel hallucination samples~\cite{zhang2021hallucination} in the missing categories, which relieves the negative impact of mismatched semantics and achieves more effective adaptation.
	
	To overcome the aforementioned challenges, we propose a SemantIc-complete Graph MAtching (SIGMA) framework for DAOD, which completes domain-mismatched semantics and reformulates the adaptation as a graph matching problem, i.e., finding the suitable matching between graph nodes to bridge the domain gap. As shown in Figure~\ref{motivation}, we design a Graph-embedded Semantic Completion module (GSC) to complete the mismatched semantics, which utilizes domain-level statistics to generate hallucination nodes in the missing categories. Then, we establish graphs to model class-conditional distributions for both domains and learn a graph-guided memory bank to improve the capacity of semantic completion in turn. Based on our  reformulation of domain adaptation, we propose a Bipartite Graph Matching adaptor (BGM) to solve the graph matching problem between the source and target graph, achieving a fine-grained domain alignment. We utilize graph nodes to learn semantic-aware node affinity and introduce graph edges in a structure-aware matching loss for the Quadratic Assignment Problem (QAP). This graph-matching-based domain alignment enables a fine-grained adaptation with well-matched semantics and relieves the biased and non-effective adaptation in existing prototype-based methods. To be summarized, our contributions are as follows.
	
	\begin{itemize}
		\item We propose a SemantIc-complete Graph MAtching (SIGMA) framework  for DAOD, which aligns the class-conditional distribution with graph matching. To the best of our knowledge, this work represents the first attempt to leverage graph matching theory to bridge the domain gap in the detection community.
		
		\item We propose a Graph-embedded Semantic Completion module (GSC) to complete mismatched semantics by generating hallucination nodes and a Bipartite Graph Matching adaptor (BGM) that
		reformulates DAOD as a graph matching problem to bridge the domain gap.
		
		\item Extensive experiments on three benchmarks demonstrate that SIGMA achieves state-of-the-art results and outperforms DAOD counterparts significantly.
		
	\end{itemize}
	
	\begin{figure*}[t]
		\begin{center}
			\includegraphics[width=0.98\linewidth]{ 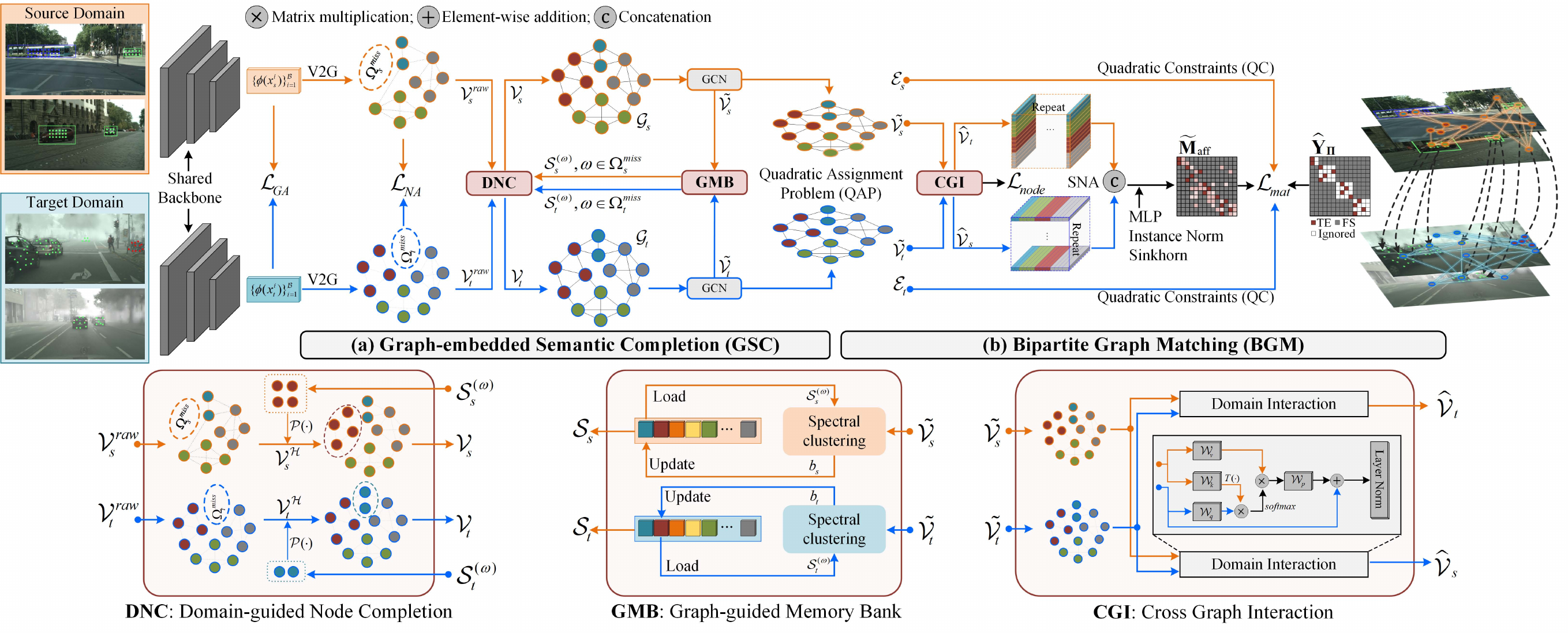}
		\end{center}
		\vspace{-15pt}
		\caption{Overview of the proposed SIGMA framework for DAOD. V2G represents vision-to-graph transformation. }
		\vspace{-10pt}
		\label{overall}
	\end{figure*}
	
	\section{Related Work}
	
	\subsection{Domain Adaptive Object Detection}
	Domain adaptive object detection (DAOD) aims to bridge the domain gap between the training and testing data, which can be categorized into style-transfer~\cite{inoue2018cross,kim2019diversify,hsu2020progressive}, self-labeling~\cite{SSAL,inoue2018cross}, and domain-alignment~\cite{DAfasterrcnn,swda,SAPN}. As one of the main streams, domain-alignment approaches adopt adversarial feature alignment and minimize the cross-domain discrepancy to bridge the domain gap. Early works align global features~\cite{DAfasterrcnn,swda,SAPN} with diverse mechanisms, e.g., spatial attention~\cite{SAPN} and strong-weak alignment~\cite{swda}. Besides, some works tend to align a community of local pixels with essential attributes, e.g., region proposals~\cite{kim2019diversify} and object centers~\cite{everypixelmatters}. Recently, some works have introduced a more precise adaptation in class-conditional distributions at the category level. GPA~\cite{xu2020crossgraph} and DBGL~\cite{bipartite_graph} explore the graph-based structural knowledge with region proposals, and model category prototypes to narrow down cross-domain measurements. PARPN~\cite{RPN} extends the idea of prototype alignment in the RPN stage, and the authors in~\cite{c2f} extend the batch-wise prototypes at the domain level. However, these works ignore the significant within-class variance, leading to a sub-optimal alignment of class-conditional distributions. This work breaks this barrier with graph matching, avoiding the inaccuracy adaptation caused by handcraft prototype design and center-based alignment.
	
	\subsection{Graph Matching}
	Graph matching establishes pair-wise node correspondences between two graphs, and gives a one-to-one matching of graph nodes belonging to different graphical entities. As a Quadratic Assignment Problem (QAP)~\cite{QAP} with combinational nature, graph matching solvers~\cite{QAP,QAP2} optimize a cross-graph permutation matrix to encode matched node pairs, considering both node and structure affinities. Recently, graph matching has been extended to visual correspondence detection~\cite{GM_VC}, multi-object tracking~\cite{GM_MOT}, point cloud registration~\cite{GM_3d} and transfer learning~\cite{GMUDA} to model pair-wise relationships in the graphical space. Gao, \etal~\cite{GM_VC} model key-point-based graphs on images and establish graph matching between images covering the same objects. Fu \etal~\cite{GM_3d} model graphs on the 3D rigid point cloud and perform graph matching on two homogeneous point sets to achieve robust point cloud registration. The authors in~\cite{GM_MOT} perform graph matching across the tracklet and detection space to achieve high-quality object tracking. Different from aforementioned scenarios with off-the-shelled graph definition and pair-wise labels, we innovatively reformulate DAOD as a graph matching problem, and leverage the QAP solver to bridge the domain gap.
	
	\section{Motivation and Preliminaries}
	We theoretically analyze existing category-level adaptation approaches, and demonstrate our motivation and new solution as follows. Considering the batch-wise source and target observation $\mathcal{S} = \{({x}_s^i, {y}_s^i)\}^{\mathcal{B}}_{i=1}$ and $\mathcal{T} = \{x_t^i\}^{\mathcal{B}}_{i=1} $ drawn from the inconsistent domain distribution $ \mathcal{P}_s$ and $ \mathcal{P}_t$ ($ \mathcal{P}_s \ne \mathcal{P}_t$), existing approaches~\cite{xu2020crossgraph,c2f,RPN,KTNet} aim to model and align class-conditional distributions $\mathcal{P}_{X|Y}(\phi(x_{s/t})|y)$, where $ \phi(\cdot)$ is the feature extractor. These works first estimate category centers $\mu^{y}_{s/t} = \mathbb{E}_{X|Y}[\phi(x)|y]$ with handcraft priors, e.g., mean-values of object features $\mu^{y}_{s/t} =\frac{1}{N_{s/t}} \sum_i^{N_{s/t}} RoI_i^{y}$, and then minimize the domain-discrepancy between $\mu^{y}_{s}$ and $\mu^{y}_{t}$. However, these methods potentially achieve a biased adaptation depending only on center-based knowledge, and fail to adapt mismatched categories $\Omega^{miss}_{s/t}$ appearing in a single domain due to the intractable $\mu^{y=\Omega^{miss}_{s/t}}_{s/t}$.

	To overcome these issues, we generate novel samples in the missing categories $\Omega^{miss}_{s/t}$ to complete the mismatched semantic, and establish a cross-image graph $\mathcal{G}_{s/t}$ to model the class-conditional distribution $\mathcal{P}_{X|Y}(\phi(x_{s/t})|y)$ for each domain. Then, we reformulate domain adaptation as a graph matching problem between $ \mathcal{G}_{s}$ and $\mathcal{G}_{t}$, which can be solved with a differential QAP~\cite{GM_3d,GM_VC,GM_MOT} as follows,
	\begin{equation}
		\vspace{-2pt}
		\begin{aligned}
			&\min_{\mathbf{\Pi}} \mathcal{F}(\mathbf{\Pi}) = ||\mathcal{A}_s - \mathbf{\Pi} \mathcal{A}_t \mathbf{\Pi}^T ||^2_{F} - tr(\mathbf{X}^T_{u}\mathbf{\Pi}),\\
			& \mathbf{\Pi} \in [0,1]^{\mathcal{N}_s \times \mathcal{N}_t}, \mathbf{\Pi} \mathbf{l}_{\mathcal{N}_s} \leq \mathbf{l}_{\mathcal{N}_t}, \mathbf{\Pi}^T \mathbf{l}_{\mathcal{N}_t} \leq  \mathbf{l}_{\mathcal{N}_s},
		\end{aligned}
		\label{eq:gm}
		\vspace{-2pt}
	\end{equation}
	where $\mathcal{A}_{s} \in \mathbb{R}^{\mathcal{N}_s \times \mathcal{N}_s}$ and $\mathcal{A}_{t} \in \mathbb{R}^{\mathcal{N}_t \times \mathcal{N}_t}$ represent the adjacent matrix encoding structure information of the graph $\mathcal{G}_{s}$ and $\mathcal{G}_{t}$ respectively, $\mathcal{N}_{s/t}$ is the number of graph nodes, $||\cdot||_{F}$ is the Frobenius norm, $\mathbf{X}_{u} \in \mathbb{R}^{\mathcal{N}_{t} \times \mathcal{N}_{s}}$ is the unary affinity matrix and generally specified as the node affinity $\mathbf{M}_{\mathrm{aff}}$~\cite{GM_VC}, and $\mathbf{\Pi}$ is the relaxed permutation matrix encoding node-to-node assignment \footnote{We follow~\cite{GM_VC} to relax the one-hot permutation matrix with continuous values to satisfy the differential requirement of neural network training.} and $\mathbf{\Pi}_{i,j}=1$ indicates that the node $v^{i}_{s} \in \mathcal{G}_{s}$ is matched with the node $v^{j}_{t} \in \mathcal{G}_{t}$.
	
	Different from existing works~\cite{xu2020crossgraph,c2f,RPN} overlooking mismatched categories, we complete missing semantics and effectively align the distribution for each appeared category. Besides, our method achieves a fine-grained adaptation guided by graph matching, breaking the barrier of existing center-based methods adopting sub-optimal alignment.

	\section{Proposed Method}
	The overall workflow the proposed SIGMA framework is shown in Figure~\ref{overall}. Given batch-wise annotated source images $\{({x}_{s}^{i}, {y}_{s}^{i})\}_{i=1}^{\mathcal{B}}$ and unlabeled target images $ \{{x}_{t}^{i}\}_{i=1}^{\mathcal{B}}$, we use a shared feature extractor $\phi$ to extract image-level features $\{\phi(x_{s/t}^i)\}_{i=1}^{\mathcal{B}}$, which are sent to Graph-embedded Semantic Completion module (GSC) (Figure~\ref{overall}(a)). In the GSC module, we first transform visual features to the graphical space (V2G) and perform domain-guided node completion (DNC) to complete mismatched semantics, obtaining semantic-complete node sets $\mathcal{V}_{s/t}$. Then, we establish cross-image graphs $\mathcal{G}_{s/t}$ to model the class-conditional distribution with enhanced nodes $\Tilde{\mathcal{V}}_{s/t}$, which also serves to learn a graph-guided memory bank (GMB) to improve the semantic completion in turn. Afterwards, the well-modeled graphs $\mathcal{G}_{s/t}$ are sent to the Bipartite Graph Matching adaptor (BGM) (Figure~\ref{overall}(b)). We use graph nodes $\Tilde{\mathcal{V}}_{s/t}$ for cross graph interaction (CGI) and learn a semantic-aware node affinity (SNA) matrix $\Tilde{\mathbf{M}}_{\mathrm{aff}}$. Besides, we leverage graph edges $\mathcal{E}_{s/t}$ to serve as quadratic constraints (QC) to optimize the graph matching permutation, achieving fine-grained adaptation with well-aligned node pairs.

	\subsection{Graph-embedded Semantic Completion}
	Given batch-wise annotated source images $ \{(x_{s}^i, y_{s}^i) \}_{i=1}^{\mathcal{B}}$ and unlabeled target images $ \{x_{t}^i\}_{i=1}^{\mathcal{B}}$ with $C$ categories, we first adopt the domain-shared backbone $\phi$ to extract visual features $ \{\phi(x_{s/t}^{i})\}_{i=1}^{\mathcal{B}}$, $\phi(x^i_{s/t}) \in \mathbb{R}^{D\times W \times H}$. For the source features, we perform spatial-uniformed sampling to collect the pixels inside ground-truth boxes as class-aware foreground nodes and a ratio $\frac{1}{C+1}$ of pixels outside foreground boxes as background samples. For the target domain, we forward-propagate target features in classification head to obtain pseudo score maps $\mathcal{M}_t \in \mathbb{R}^{C \times W \times H}$ as the surrogate sampling principle. Then we sample the pixels satisfying $\max_{C}(\mathcal{M}_{t}^{i}) > \tau_{fg}$ as class-aware foreground nodes and a ratio $\frac{1}{C+1}$ of low-score pixels ($\max_{C}(\mathcal{M}_{t}^{i}) < \tau_{bg}$) as background samples\footnote{$\tau_{fg}$ is empirically set 0.5 to satisfy the active condition of the non-linear $sigmoid$ function and $\tau_{bg}$ is set 0.05 following the commonly used score-threshold setting in existing object detectors~\cite{fasterrcnn,retinanet,fcos,yolov3}.}. After sampling fine-grained visual features, we perform a non-linear projection to obtain the raw node embedding $\mathcal{V}^{raw}_{s/t} =\{v_{s/t}^{i}\}_{i=1}^{\mathcal{N}_{s/t}}$, achieving the transformation from the visual space to the graphical space.

	\noindent\textbf{Domain-guided Node Completion.} The object categories $\Omega_{s/t}^{\mathcal{B}} \in \{0, 1,..., C\}$ within a training batch are always mismatched between the source and target domain, limiting the adaptation of class-conditional distributions. Hence, we propose a semantic completion strategy to generate hallucination nodes in missing categories $\Omega_s^{miss}= \{ \omega| \omega\in \Omega^{\mathcal{B}}_{t}, \omega \notin \Omega^{\mathcal{B}}_{s} \}$, $\Omega_{t}^{miss}= \{ \omega| \omega\in \Omega^{\mathcal{B}}_{s}, \omega \notin \Omega^{\mathcal{B}}_{t} \}$, obtaining semantic-complete nodes $\mathcal{V}_{s/t}$. To generate additional nodes containing non-existing semantics, we define a graph-guided memory bank $\mathcal{S}_{s/t} \in \mathbb{R}^{C \times D}$ to save the category-specific knowledge of inner-domain semantics, and we will explain the learning strategy of this memory bank in the next section. Considering the source and target domains share a similar category space~\cite{DAfasterrcnn}, we fully utilize the semantic cues from the counterpart domain to guide the node generation, which provide a joint measurement of the class-conditional distribution within the batch. Specifically for the completion of the source-missing category $\omega \in \Omega_s^{miss}$, we calculate the standard variance of target nodes $\{v_{t}^{(\omega)}\}$ in class $\omega$ to obtain a variant vector $\sigma_{t}^{(\omega)} \in \mathbb{R}^{D}$, which approximates the scale of the distribution for the missing category $\omega$. Then, we load the corresponding memory seed $\mathcal{S}_{s}^{(\omega)}$ from the memory bank to serve as the category-specific expectation $\mu_{s}^{(\omega)}$. After that, we perform Gaussian sampling and adopt a linear projection $\mathcal{P}(\cdot)$ to obtain hallucination nodes ${\mathcal{V}}^{\mathcal{H}}_{s} = \{v_s^{h}| v_s^{h}=\mathcal{P}(x_s^{h}), x_s^{h} \sim N(\mu^{(\omega)}_s, \sigma^{(\omega)}_t ) \}$ belonging to the mismatched categories. The same completion is also conducted in the target domain to obtain the nodes ${\mathcal{V}}^{\mathcal{H}}_{t}$ in the target-missing categories $\Omega_{t}^{miss}$. Instead of aligning these statistic-based estimations directly~\cite{xu2020crossgraph,c2f,RPN}, we fully utilize domain knowledge to generate novel and unbiased samples, avoiding the biased and sub-optimal alignment. Finally, both existing nodes and hallucination ones constitute the semantic-complete node set ${\mathcal{V}}_{s/t}$ for the followed graph modelling.

	\noindent\textbf{Graph-guided Memory Bank.} Since the nodes ${\mathcal{V}}_{s/t}$ derive from different images within a batch, we establish a cross-image graph to model the class-conditional distribution with long-distance semantic dependency, and propose a memory bank to preserve graph-based knowledge, which helps the DNC to generate better hallucination nodes in turn. Specifically, we first introduce edge connections $\mathcal{E}_{s/t}$ between nodes $\mathcal{V}_{s/t}$ and set up a cross-image graph $\mathcal{G}_{s/t}=\{\mathcal{V}_{s/t},\mathcal{E}_{s/t}\}$ in each domain. For the graph edge, we utilize edge drop~\cite{dropedge} to avoid the potential relationship bias caused by the abundant visual correspondence: $\mathcal{A}_{s/t} = Edgedrop\{softmax[\mathcal{V}_{s/t} \mathcal{W}_{e} (\mathcal{V}_{s/t} \mathcal{W}_{e})^{T}]\}$, where $\mathcal{A}_{s/t}$ is the adjacent matrix encoding structure information, and $\mathcal{W}_{e}$ is a learnable linear projection. Then, we perform single-layer graph convolution with the graph-based message propagation among nodes to aggregate cross-image semantic knowledge, yielding the enhanced node representation: $\Tilde{v}^i_{s/t} = \mathrm{LN}(\sum^{|{\mathcal{NR}}^{i}|}_{v^j_{s/t} \in \mathcal{NR}^i} \mathcal{A}^{i,j}_{s/t} {v}_{s/t}^{j} \mathcal{W}_{gcn} + {v}_{s/t}^i)$, where $\mathcal{NR}^i$ represents the neighbour nodes of $v^i_{s/t}$, $\mathcal{W}_{gcn}$ is the learnbale parameter, and $\mathrm{LN}$ is the layer normalization~\cite{LN}.
	
	To provide representative and robust dependency for the hallucination node generation, we introduce a memory bank to save class-specific graph embedding and design a cluster-based update strategy for the memory bank learning. Specifically, we randomly initialize a memory bank $\mathcal{S}_{s/t} \in \mathbb{R}^{C\times D}$ at the beginning of the training and gradually update memory seeds with appeared graph nodes. For each appeared category $\omega$ within a training batch, we collect graph nodes $\{\Tilde{v}^{(\omega)}_{s/t}\},\Tilde{v}^{(\omega)}_{s/t} \in \mathbb{R}^{D} $ in class $\omega$ and load the corresponding memory seed $\mathcal{S}_{s/t}^{(\omega)} \in \mathbb{R}^{D}$ from the memory bank $\mathcal{S}_{s/t}$. Then, we get both the memory seed and graph nodes together $\{\mathcal{S}_{s/t}^{(\omega)}, \Tilde{v}^{(\omega)}_{s/t}\}$ and conduct spectral clustering~\cite{cluster} in the graphical space to generate two clusters, i.e., a seed-included cluster ${\pi}_{s/t}^{seed}=\{\mathcal{S}_{s/t}^{(\omega)}, \Tilde{v}^{(\omega)}_{s/t}\}$ and an ``else" cluster $\pi^{else}=\{ \Tilde{v}^{(\omega)}_{s/t}\}$. Since the domain-level knowledge, referred to as the memory seed, provides a more robust and precise estimation compared with the batch-wise measurement, we only utilize the nodes in ${\pi}_{s/t}^{seed}$ to update the memory bank, which relieves the impact of noisy nodes appeared in the early training stage:
	\begin{equation}
		\mathcal{S}_{s/t}^{(\omega)} \gets sim(b_{s/t}, \mathcal{S}^{(\omega)}_{s/t}) \mathcal{S}^{(\omega)}_{s/t} +[1- sim(b_{s/t}, \mathcal{S}^{(\omega)}_{s/t})] b_{s/t},
	\end{equation}
	where $sim(b_{s/t}, \mathcal{S}^{(\omega)}_{s/t})=\frac{b_{s/t} \cdot \mathcal{S}^{(\omega)}_{s/t}}{\left\|b_{s/t}\right\|_{2} \cdot\left\| \mathcal{S}^{(\omega)}_{s/t}\right\|_{2}}$ indicates the adaptive momentum for better gradient-free learning~\cite{c2f,Mega-DA}, and $b_{s/t}=\frac{1}{|\pi^{seed}_{s/t}|-1}\sum_{\Tilde{v}^{(\omega)}_{s/t} \in \pi^{seed}_{s/t}} \Tilde{v}^{(\omega)}_{s/t}$. We only utilize existing graph nodes to update memory seeds, and remove those hallucination ones to avoid the potential negative impact of handcraft Gaussian priors for the model learning.
	
	\subsection{Bipartite Graph Matching}
	Given the graph $\mathcal{G}_{s/t}$, we reformulate the cross-domain alignment as a graph matching problem, i.e., solving the QAP between $\mathcal{G}_s$ and $\mathcal{G}_t$. Specifically, we use graph nodes $\Tilde{\mathcal{V}}_{s/t}$ to establish cross-graph interaction and learn a node affinity $\Tilde{\mathbf{M}}_{\mathrm{aff}}$. Besides, we introduce graph edges $\mathcal{E}_{s/t}$ to bridge the domain gap with a structure-aware matching loss.
	
	\noindent\textbf{Cross Graph Interaction.} Since graph matching is a collaborative optimization problem between two graphical entities, the message propagation across graphs is essential for the optimal solution in graph-based affinity learning. Hence, we introduce the knowledge exchange between $\mathcal{G}_{s}$ and $\mathcal{G}_{t}$ to establish the cross-domain semantic interaction:
	\begin{equation}
		\begin{aligned}
			&\hat{\mathcal{V}}_{s} = \mathrm{LN}\{softmax[(\Tilde{\mathcal{V}}_{s}\mathcal{W}_{q}) (\Tilde{\mathcal{V}}_{t}\mathcal{W}_{k})^{T}] (\Tilde{\mathcal{V}}_{t} \mathcal{W}_{v})\mathcal{W}_{p} + \Tilde{\mathcal{V}}_{s}\}, \\
			&\hat{\mathcal{V}}_{t} = \mathrm{LN}\{softmax[(\Tilde{\mathcal{V}}_{t}\mathcal{W}_q) (\Tilde{\mathcal{V}}_{s}\mathcal{W}_k)^{T}] (\Tilde{\mathcal{V}}_{s} \mathcal{W}_{v})\mathcal{W}_{p}+ \Tilde{\mathcal{V}}_{t}\},
		\end{aligned}
		\label{eq:interaction}
	\end{equation}
	where $\hat{\mathcal{V}}_{s/t} =\{\hat{v}^i_{s/t}\}_{i=1}^{\mathcal{N}_{s/t}} $ is the graph node set with cross-domain perception, $\mathrm{LN}$ is the layer normalization~\cite{LN}, and $\mathcal{W}_{(\cdot)}$ are learnable parameters. To enhance the graphical semantics, we introduce an auxiliary node classification task by adopting a classifier $f_{cls}$ with the Cross Entropy loss:
	\begin{equation}
		\mathcal{L}_{node} =-\sum_{i=1}^{\mathcal{N}_{s}+\mathcal{N}_{t}} {y_i log \{softmax[f_{cls}(\hat{v}^i_{s/t})]\}},
		\label{eq:node}
	\end{equation}
	where $y_i$ represents the ground-truth label for source nodes and the pseudo label (obtained from score maps $\mathcal{M}_t$) for target nodes. Dense relationships can be established among nodes belonging to different domains, serving the sparse and fine-grained adaptation with interactive semantic cues.

	\noindent\textbf{Semantic-aware Node Affinity.} Given the graph nodes $\hat{\mathcal{V}}_{s/t}$ with cross-domain perception, we further learn an affinity matrix to model the node correspondence between $\mathcal{G}_{s}$ and $\mathcal{G}_{t}$. Different from existing graph matching approaches~\cite{GM_3d,GM_VC,GM_MOT} utilizing local visual representations, we leverage the category-level semantic with inherent relationships to learn a semantic-aware affinity matrix. Specifically, we define the entry of the node affinity matrix as follows: $ \mathbf{M}_{\rm{aff}}^{i,j} = f_{mlp}\{f_{p} (\hat{v}^i_s) \stackMath\mathbin{\stackinset{c}{0ex}{c}{0ex}{\mathrm{c}}{\bigcirc}} f_{p}(\hat{v}_t^j)\}$, $\mathbf{M}_{\rm{aff}} \in \mathbb{R}^{\mathcal{N}_{s} \times \mathcal{N}_t}$, where $\stackMath\mathbin{\stackinset{c}{0ex}{c}{0ex}{\mathrm{c}}{\bigcirc}}$ is the concatenation operation, $f_{p}$ indicates a linear projection, and $f_{mlp}$ is a multi-layer perceptron layer (MLP) with a single output channel. This MLP layer learns inherent semantic relationships between two graph nodes and encodes them into affinity representations. $\mathbf{M}_{\rm{aff}}$ is then sent to the Instance Normalization layer as~\cite{GM_3d} and the differential Sinkhorn layer~\cite{Sinkhorn1964ARB} to obtain a double-stochastic affinity matrix $\Tilde{\mathbf{M}}_{\mathrm{aff}}$ with maximum $k$-iteration optimization (k is set 20 enough for optimization). Finally, each positive entry in the affinity matrix $\Tilde{\mathbf{M}}_{\mathrm{aff}}$ indicates a matched node pair across two graphs for fine-grained domain adaptation.
	
	\newcolumntype{g}{>{\columncolor{Gray}} p{0.55cm}}
	\begin{table*}
		\begin{center}
			\small
			\begin{tabular}{l |c |cccccccc| >{\columncolor{Gray}}c >{\columncolor{Gray}}c}
				\toprule
				Method &Backbone & person &rider& car &truck &bus &train &motor &bike & mAP & SO/ GAIN \\
				\midrule
				CFFA~\cite{c2f}$_{CVPR'20}$ &\multirow{10}*{{VGG-16}} &34.0 &46.9& 52.1 &30.8& 43.2& 29.9& 34.7& 37.4& 38.6& 20.8/ 17.8 \\
				EPM \cite{everypixelmatters}$_{ECCV'20}$&& 41.9 &38.7 &56.7 &22.6 &41.5 &26.8 &24.6 &35.5 &36.0 &18.4/ 17.6 \\
				RPNPA~\cite{RPN}$_{CVPR'21}$&&33.6& 43.8&49.6&{32.9}&45.5&46.0&35.7&36.8&40.5& 20.8/ 19.7   \\
				
				UMT~\cite{UMT}$_{CVPR'21}$&&33.0&46.7&48.6&\textbf{34.1}&\textbf{56.5}&{46.8}&30.4&37.4 & 41.7&21.8/ 19.9  \\
				MeGA~\cite{Mega-DA}$_{CVPR'21}$&& 37.7& \textbf{49.0} &52.4 &25.4& 49.2 &\textbf{46.9}&{34.5} &39.0&41.8&24.4/ 17.4    \\
				
				ICCR-VDD~\cite{ICCD}$_{ICCV'21}$& &33.4& 44.0& 51.7& 33.9& {52.0}& 34.7& 34.2& 36.8& 40.0& 22.8/ 17.2  \\
				KTNet~\cite{KTNet}$_{ICCV'21}$& &46.4 &43.2& {60.6}& 25.8& 41.2& 40.4& 30.7& 38.8& 40.9&18.4/ 22.5 \\
				SSAL~\cite{SSAL}$_{NeurIPS'21}$& & 45.1 & 47.4 & 59.4 & 24.5 & 50.0  &25.7 & 26.0 & 38.7 & 39.6 & 20.4/ 19.2   \\
				
				SIGMA (ours)&&  \textbf{46.9}&{48.4}&\textbf{63.7}&27.1& 50.7&35.9&{34.7}&\textbf{41.4}&\textbf{43.5} &18.4/ \textbf{25.1}\\
				
				\midrule
				GPA~\cite{xu2020crossgraph}$_{CVPR'20}$& \multirow{6}*{{ResNet-50}}& 32.9& 46.7 &54.1 &24.7 &45.7& 41.1& 32.4& 38.7& 39.5&  22.8/ 16.7 \\
				EPM~\cite{everypixelmatters}$_{ECCV'20}$&&39.9 &38.1 &57.3 &28.7 &50.7 &37.2& 30.2& 34.2& 39.5 &24.2/ 15.3 \\
				DIDN~\cite{DIDN}$_{ICCV'21}$& &38.3& 44.4& 51.8& 28.7 &\textbf{53.3}& 34.7& 32.4& 40.4& 40.5& 28.6/ 11.9  \\
				DSS~\cite{DSS}$_{CVPR'21}$&&42.9& \textbf{51.2}& 53.6& \textbf{33.6}& 49.2 &18.9&\textbf{36.2}& \textbf{41.8}& 40.9 & 22.8/ 18.1  \\
				SDA~\cite{SDA}$_{ICCV'21}$& &38.8 &45.9& 57.2& 29.9& 50.2& \textbf{51.9}& 31.9& 40.9& 43.3& 22.8/ \textbf{20.5}  \\
				
				SIGMA (ours) &&\textbf{44.0}&43.9&\textbf{60.3}&31.6&50.4&51.5&31.7&40.6&\textbf{44.2} &24.2/ 20.0 \\
				
				\bottomrule
			\end{tabular}
		\end{center}
		\vspace{-12pt}
		\caption{Results on Cityscapes$\to$Foggy Cityscapes (\%) with VGG-16 and ResNet-50 backbone networks. SO represents the source only results and GAIN indicates the adaptation gains compared with the source only model.}
		\label{tab:c2f}
		\vspace{-5pt}
	\end{table*}
	
	\noindent\textbf{Structure-aware Matching Loss.} Since graph nodes are drawn from the graphically modeled class-conditional distribution, we align the node pairs across two domains with homogeneous semantics ($\hat{v}_s^{(\omega)} \in \mathcal{G}_s$ and $\hat{v}_t^{(\omega)} \in \mathcal{G}_t$), to adapt the distribution for category $\omega$. Specifically, we propose a structure-aware matching loss to achieve this fine-grained domain adaptation with node-to-node graph matching, which consists of three components as follows,
	\begin{equation}
		\begin{aligned}
			\mathcal{L}_{mat} & = \sum_{i}\frac{1}{\mathcal{N}_s} [\max_{j}(\Tilde{\mathbf{M}}_{\rm{aff}}\odot \mathbf{Y}_{\mathbf{\Pi}})_{i,j}- \mathbf{1} ]^{2} \\
			& + \sum_{i,j} \frac{1}{||\mathbf{1}-\mathbf{Y}_{\mathbf{\Pi}}||_{1}}    [\Tilde{\mathbf{M}}_{\rm{aff}}\odot (\mathbf{1} - \mathbf{Y}_{\mathbf{\Pi}})]^{2}_{i,j} \\
			& + \sum_{i,j} \frac{1}{\mathcal{N}_{s}\cdot \mathcal{N}_{t}} (\mathcal{A}_{s}\Tilde{\mathbf{M}}_{\rm{aff}} - \Tilde{\mathbf{M}}_{\rm{aff}} \mathcal{A}_t)_{i,j},
		\end{aligned}
		\label{eq:match}
		\vspace{-4pt}
	\end{equation}
	where the $(i,j)$ entry in $\mathbf{Y}_{\mathbf{\Pi}}\in \mathbb{R}^{\mathcal{N}_s \times \mathcal{N}_t}$ is $\mathbf{1}$ if $v^i_s \in \mathcal{G}_s$ and $v^j_t\in \mathcal{G}_t$ are in the same category $\omega$, otherwise $\mathbf{0}$, and $\Tilde{\mathbf{M}}_{\mathrm{aff}} \in \mathbb{R}^{\mathcal{N}_s \times \mathcal{N}_t}$ is the node affinity. The first term works on correctly matched node pairs and enhances the \textit{best-matching} of correct cases, named True-positive Enhancement (TE) (as the \textit{Red} entries of Figure~\ref{overall} $\hat{\mathbf{Y}}_{\mathbf{\Pi}}$ ). The second term evaluates the difference between the node affinity and ground-truth to suppress wrongly activated cases, i.e., False-positive Suppression (FS) (as the \textit{Grey} entries of Figure~\ref{overall} $\hat{\mathbf{Y}}_{\mathbf{\Pi}}$). Besides, we introduce structure-aware Quadratic Constrains (QC) as the third term to minimize the structural difference of matched node pairs in a local neighborhood. Based on the consistent objective of Eq.~\ref{eq:gm} and Eq.~\ref{eq:match} about graph matching, each source node will be aligned to the optimal-matched counterpart in the target domain in the same category, achieving a fine-grained alignment of class-conditional distributions during training.
	
	\subsection{Model Optimization}
	
	During training, we adopt class-agnostic global alignment~\cite{everypixelmatters} on visual features $ \{{x}_{s/t}^{i}\}_{i=1}^{\mathcal{B}}$ with adversarial loss $\mathcal{L}_{GA}$. Considering the non-grid correspondence among graph nodes and the non-euclidean representation of graphical space~\cite{graph_da}, we design a Node Discriminator (ND) to align well-matched nodes, consisting a gradient reversed layer~\cite{grl}, three stacked discrimination blocks $f_{b}$ (each block is FC-LayerNrom-ReLU), and a domain classifier $f_{dc}$ followed with the Binary Cross Entropy (BCE) loss:  $\mathcal{L}_{NA}=-\sum_i^{\mathcal{N}_s}\mathcal{D}log\{f_{dc}[f_{b}(v_{s}^i)]\}-\sum_{i}^{\mathcal{N}_t}(1-\mathcal{D})log\{f_{dc}[f_{b}(v_{t}^i)]\}$, where $\mathcal{D}$ is the domain label as~\cite{DAfasterrcnn} and $v^i_{s/t}$ are existing graph nodes. Then, the overall optimization objective of the proposed framework is denoted as:
	\begin{equation}
		\begin{aligned}
			\mathcal{L}= \lambda_{1}\mathcal{L}_{node} + \lambda_{2}\mathcal{L}_{mat} +\mathcal{L}_{NA}+ \mathcal{L}_{GA} + \mathcal{L}_{det},
		\end{aligned}
	\end{equation}
	where $\mathcal{L}_{node}$ is the node classification loss, $\mathcal{L}_{mat}$ is the graph matching loss, $\mathcal{L}_{NA}$ is the node alignment loss, $\mathcal{L}_{GA}$ is the global alignment loss~\cite{everypixelmatters} and $\mathcal{L}_{det}$ is the detection loss. $\lambda_{1/2}$ are set 0.1 respectively to control the intensity.
	
	\begin{table}[t]
		\begin{center}
			\footnotesize
			\begin{tabular}{p{2.5cm}| >{\columncolor{Gray}}p{0.65cm}<{\centering} >{\columncolor{Gray}}p{1.2cm}<{\centering}|>{\columncolor{Gray}}p{0.65cm}<{\centering} >{\columncolor{Gray}}p{1.2cm}<{\centering} }
				\toprule
				Method & S$\to$C& SO/GAIN& K$\to$C &SO/GAIN\\
				\midrule
				EPM~\cite{everypixelmatters}$_{ECCV'20}$        & 49.0 & 39.8/ 9.2 & 43.2 &34.4/ 8.8 \\
				DSS~\cite{DSS}$_{CVPR'21}$                      & 44.5 &34.7/ 9.8 & 42.7 &34.6/ 8.1\\
				MEGA~\cite{Mega-DA}$_{CVPR'21}$                 & 44.8 & 34.3/ 10.5& 43.0 &30.2/ \textbf{12.8}\\
				RPNPA~\cite{RPN}$_{CVPR'21}$                    & 45.7 & 34.6/ 11.1& -&-\\
				UMT~\cite{UMT}$_{CVPR'21}$                      & 43.1 &34.3/ 8.8 & -&-\\
				KTNet~\cite{KTNet}$_{ICCV'21}$                  & 50.7 &39.8/ 10.9& 45.6 &34.4/ 11.2\\
				SSAL~\cite{SSAL}$_{NeurIPS'21}$                    & 51.8 &38.0/ 13.8& 45.6 &34.9/ 10.7\\
				SIGMA (ours) & \textbf{53.7}  & 39.8/ \textbf{13.9}& \textbf{45.8} &34.4/ 11.4\\
				\bottomrule
				
			\end{tabular}
		\end{center}
		\vspace{-12pt}
		\caption{Comparison results (\%) on Sim10K$\to$Cityscapes (S$\to$C) and KITTI$\to$Cityscapes (K$\to$C) with VGG-16 backbone. }
		\label{tab: sim10k}
		\vspace{-5pt}
	\end{table}
	
	\section{Experiments}
	
	\subsection{Datasets and Evaluation}
	
	We conduct extensive experiments on three adaptation scenarios following the standard UDA setting in existing literature~\cite{DAfasterrcnn,everypixelmatters,KTNet,SSAL}. We use the mean Average Precision with different IoU thresholds (mAP$_{IoU}$) for comparison and utilize SO/GAIN to assess the source only results\footnote{Source Only (SO) indicates training with labeled source images and testing on the target data, which is the same as ``w/o adapt".} and the adaptation gains compared with the SO. Besides, we also report the results of GA~\cite{everypixelmatters} that adopts global alignment~\cite{DAfasterrcnn} on the FCOS~\cite{fcos} detector as our baseline counterpart.
	
	\noindent\textbf{Cityscapes$\to$Foggy Cityscapes.} The Cityscapes~\cite{cordts2016cityscapes} is a street scene datasets captured with on-board cameras under the dry weather condition, which consists of the \textit{train} set (2975 images) and \textit{validation} set (500 images) with eight categories of annotated bounding boxes. Foggy Cityscapes~\cite{sakaridis2018foggy} is a synthesized dataset based on the Cityscapes with foggy noise. We explore the weather conditioned domain gap in this adaptation scenario.
	
	\noindent\textbf{Sim10k$\to$Cityscapes.} Sim10k~\cite{johnson2017sim10k} is a simulated dataset obtained from the video game Grand Theft Auto V, yielding the domain gap with the real-world scene (Cityscapes). This dataset covers 10,000 images of the annotated bounding boxes in the car category. We perform domain adaptation between synthesized and real-world images and report the performance on car category as the common setting.
	
	\noindent\textbf{KITTI$\to$Cityscapes.} KITTI~\cite{geiger2012kitti} is a real-world traffic scene dataset collected from vehicle-mounted cameras, which yields the cross-camera domain gap with Cityscapes (on-board cameras). This dataset covers annotated cars in 7,481 images with cross-camera domain gap for adaptation.

	\subsection{Implementation Details}
	We adopt both VGG-16~\cite{vgg} and ResNet-50~\cite{resnet} feature extractors, which are implemented with Pytorch~\cite{paszke2019pytorch}. Our model is trained with the Stochastic Gradient Descent (SGD) optimizer with a 0.0025 learning rate, 4 batch-size, momentum of 0.9, and weight decay of 5$\times$10$^{-4}$. We sample at most 100 graph nodes for each feature map in each domain. Considering the graph matching may fail if no nodes appear in the target domain, we follow~\cite{everypixelmatters} to pretrain the framework as a warm-up stage before introducing the BGM adaptor. The adaption-unrelated settings about the object detector strictly follow related works~\cite{everypixelmatters,KTNet,SSAL}.
	
	\subsection{Comparison with State-of-the-arts}
	
	\noindent\textbf{Cityscapes$\to$Foggy Cityscapes.} We present the comparison with VGG-16 and ResNet-50 backbones in Table~\ref{tab:c2f}. SIGMA achieves 43.5\% and 44.2\% mAP, respectively, outperforming existing works by a large margin. Compared with category-level adaptation approaches, e.g., CFFA~\cite{c2f} (38.6\%), RPNPA~\cite{RPN} (40.5\%), MeGA-CDA~\cite{Mega-DA} (41.8\%), KTNet~\cite{KTNet} (40.9\%), and GPA~\cite{xu2020crossgraph} (39.5\%), SIGMA achieves 4.9\%, 3.0\%, 1.7\%, 2.6\%, and 4.7\% mAP improvements respectively, showing our advantages over existing prototype-based works. Besides, SIGMA surpasses EPM~\cite{everypixelmatters}, KTNet~\cite{KTNet}, and SSAL~\cite{SSAL} with 7.5\%, 2.6\%, and 3.9\% mAP using the same FCOS~\cite{fcos} object detector.
	
	\noindent\textbf{Sim10k$\to$Cityscapes.} The experimental comparison is recorded in the left part of Table~\ref{tab: sim10k}. SIGMA achieves a 53.7\% mAP with the best adaptation gain (13.9\% AP), outperforming existing works significantly. Compared with the approaches using the same FCOS~\cite{fcos} object detector, e.g., EPM~\cite{everypixelmatters} (49.0\% mAP), KTNet~\cite{KTNet} (50.7\% mAP), SSAL~\cite{SSAL} (51.8\% mAP),  SIGMA gives 4.7\%, 3.0\%, and 1.9\% mAP improvements, verifying our effectiveness.
	
	\noindent\textbf{KITTI$\to$Cityscapes.} The comparison results are shown in the right part of Table~\ref{tab: sim10k}. SIGMA outperforms existing works with a 45.8\% mAP and achieves a comparable adaptation gain (11.4\% mAP) compared with state-of-the-arts. Compared with EPM~\cite{everypixelmatters}, KTNet~\cite{KTNet} and SSAL~\cite{SSAL}, our method shows the advantage in terms of adaptation.
	
	\begin{table}[t]
		\footnotesize
		\begin{center}
			\begin{tabular}{ p{0.8cm} |p{0.6cm}<{\centering}  |p{0.2cm}<{\centering}  p{0.2cm}<{\centering}  p{0.2cm}<{\centering}  p{0.2cm}<{\centering}  p{0.15cm}<{\centering}  p{0.2cm}<{\centering}  p{0.2cm}<{\raggedleft}
					p{0.35cm}<{\centering} |g}
				\toprule
				Method & w/o & prsn &rider& car &truc &bus &train &moto &bike & mAP\\
				\midrule
				GA~\cite{everypixelmatters}&-& 40.3&41.5&54.2&26.7&42.1&15.4&27.1&35.1 &35.3 \\
				\midrule
				\multirow{4}*{+GSC}& DNC &45.2&46.2&57.2&29.1&46.5&31.2&29.2&38.7&40.4  \\
				& GMB &43.5&43.8&57.4&\textbf{29.4}&48.3&30.4&31.4&41.1&41.0  \\
				& ND& 44.1&45.2&56.7&28.0&45.9&23.9&32.8&38.7&39.4  \\
				& - & 45.8&47.6&58.9&27.3&48.6&33.8&32.7&39.3&41.8  \\
				\midrule
				& CGI& 44.4&48.0&58.8&28.4&50.3&\textbf{40.5}&31.7&40.8&42.8  \\
				{+GSC}& SNA&46.0&46.9&58.8&28.6&48.2&40.4&33.1&39.5&42.6  \\
				{+BGM} & SML&46.1&\textbf{49.9}&59.1&26.2&\textbf{52.5}&27.1&34.6&41.3&42.2  \\
				
				& -&\textbf{46.9}&48.4&\textbf{63.7}&27.1&50.7&35.9&\textbf{34.7}&\textbf{41.4}&\textbf{43.5}\\
				
				\bottomrule
			\end{tabular}
		\end{center}
		\vspace{-10pt}
		\caption{Ablation studies on Cityscapes$\to$Foggy Cityscapes (\%).}
		\label{tab: ablation}
		\vspace{-5pt}
	\end{table}
	
	\begin{table}
		\footnotesize
		\begin{center}
			\begin{tabular}{p{0.5cm}<{\centering} |p{0.5cm}<{\centering}  |p{0.25cm}<{\centering}  p{0.25cm}<{\centering}  p{0.25cm}<{\centering}  p{0.25cm}<{\centering}  p{0.15cm}<{\centering}  p{0.25cm}<{\centering}  p{0.25cm}<{\centering}
					p{0.35cm}<{\centering}|g}
				\toprule
				$\mathcal{N}^{f}_s$ &$\mathcal{N}^{f}_t$ & prsn &rider& car &truc &bus &train &moto &bike &mAP\\
				\midrule
				200&0&41.2&45.1&55.2&26.9&44.2&16.3&28.9&37.0  &36.8\\
				0&200 &42.4&41.8&55.3&27.7&44.0&21.8&29.2&36.6&37.3 \\
				\midrule
				20&20&42.4&44.0&56.5&27.3&45.8&26.6&30.9&38.6&39.0\\
				50&50&44.2&43.4&56.9&\textbf{32.2}&45.7&38.6&29.6&37.5&41.0\\
				100&100&\textbf{46.9}&48.4&\textbf{63.7}&27.1&50.7&35.9&\textbf{34.7}&\textbf{41.4}&43.5 \\
				200&200&44.3&\textbf{48.8}&59.0&28.9&{51.7}&\textbf{45.1}&34.2&39.9&\textbf{43.9}  \\
				\midrule
				500&500&44.4&47.1&58.0&24.4&\textbf{52.5}&40.3&31.2&40.1&42.6  \\
				\bottomrule
			\end{tabular}
		\end{center}
		\vspace{-10pt}
		\caption{Results on Cityscapes$\to$Foggy Cityscapes (\%) with different node combinations. $\mathcal{N}^{f}_{s/t}$ represent the maximum sampled nodes from source and target domains in each feature map.}
		\label{tab: node}
		\vspace{-5pt}
	\end{table}
	
	\begin{table}
		\footnotesize
		\begin{center}
			\begin{tabular}{c |l| c  c c}
				\toprule
				Strategy & Loss& mAP$_{0.5:0.95}$& mAP$_{0.5}$ &mAP$_{0.75}$\\
				\midrule
				\multirow{2}{*}{Single}
				&+TE&22.0&42.1&20.3\\
				\multirow{2}{*}{matching}&+TE+FS&23.8&43.2&23.0\\
				&+TE+FS+QC&\textbf{24.0}&\textbf{43.5}&\textbf{23.5}\\
				\midrule
				Multiple&+BCE&23.2& 42.9&22.8\\
				matching&+MSE&23.7&43.1&23.0\\
				\bottomrule
			\end{tabular}
		\end{center}
		\vspace{-10pt}
		\caption{Results on Cityscapes$\to$Foggy Cityscapes (\%) with different matching strategies and loss functions. mAP$_{0.5:0.95}$ is the averaged mAP from 0.5 to 0.95 IoU with 0.05 intervals. BCE is Binary Cross Entropy and MSE is Mean Squared Error.}
		\label{tab: match}
		\vspace{-15pt}
	\end{table}
	
	\begin{figure*}[t]
		\begin{center}
			\includegraphics[width=0.93\linewidth]{ 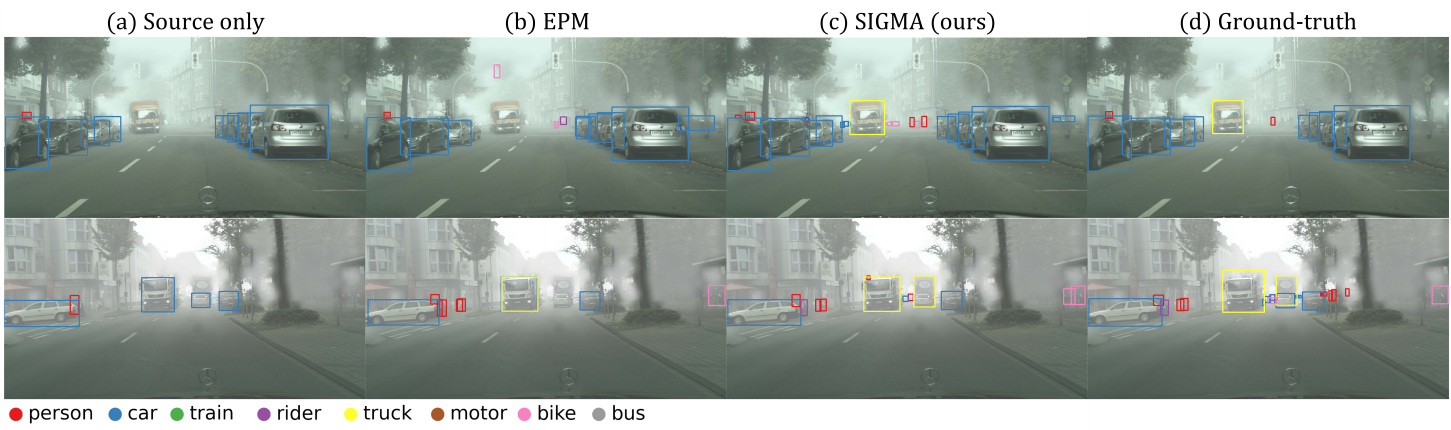}
		\end{center}
		\vspace{-15pt}
		\caption{Result comparison on the {Cityscapes$\to$Foggy Cityscapes} adaptation scenario among (a) the source only model, (b) EPM~\cite{everypixelmatters}, (c) the proposed SIGMA, and (d) Ground-truth. (Zooming in for best view.)}
		\vspace{-15pt}
		\label{show}
	\end{figure*}
	
	\begin{figure}[t]
		\begin{center}
			\includegraphics[width=0.85\linewidth]{ 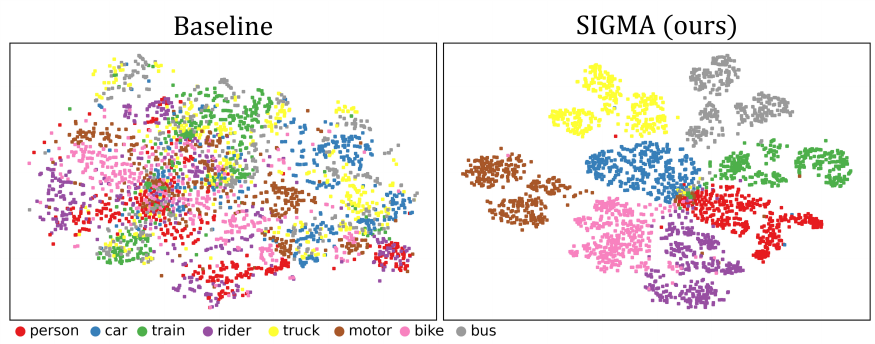}
		\end{center}
		\vspace{-15pt}
		\caption{Feature comparison via T-SNE between the baseline model and our method. For each category, we randomly sample object features (marked as squares) inside bounding boxes in the source domain and target domain equally. }
		\vspace{-10pt}
		\label{tsne}
		
	\end{figure}
	
	\subsection{Ablation Studies}
	
	We report detailed ablation studies (Table~\ref{tab: ablation}) conducted on Cityscapes$\to$Foggy Cityscapes with VGG-16 backbone.
	
	\noindent\textbf{Graph-embedded Semantic Completion.} As shown in Table~\ref{tab: ablation}, adopting the GSC module can achieve 41.8\% mAP with 6.5\% mAP gains compared with the GA baseline~\cite{everypixelmatters}. We then gradually remove each sub-component to verify its effectiveness. Removing Domain-guided Node Completion (DNC) limits the model optimization under mismatched semantic knowledge (40.4\% mAP). Replacing the Graph-guided Memory Bank (GMB) with a common buffer gives 0.8\% mAP drops (41.0\% mAP) due to the impact of unavoidable noisy samples, and removing Node Discriminator (ND) gives a significant drop (39.4\%) due to the severe domain gap in the graphical space.
	
	\noindent\textbf{Bipartite Graph Matching.} Introducing the BGM adaptor achieves consistent improvements with a remarkable 43.5\% mAP, outperforming the baseline model with 8.2\% mAP. Removing Cross Graph Interaction (CGI) gives a 0.7\% mAP performance drop (42.8 \% mAP) due to the limited interaction between two domains. Replacing the Semantic-aware Node Affinity (SNA) with the simplified strategy in~\cite{GM_VC} leads to 0.9\% mAP drops (42.6\% mAP), and removing the Structure-aware Matching Loss (SML) reduces the performance (42.1\% mAP). Hence, each sub-component is necessary for SIGMA to achieve state-of-the-art results.
	
	\subsection{Sensitivity Analysis}
	
	To better understand our method, we investigate the node selection (Table~\ref{tab: node}) and matching design (Table~\ref{tab: match}).
	
	\noindent\textbf{Evaluation on the number of nodes.} As shown in Table~\ref{tab: node}, we compare different node combinations ($\mathcal{N}^{f}_{s/t}$ represents the maximum number of nodes sampled from each feature map). Only utilizing source and target nodes ($1^{st}$ and $2^{nd}$ lines) severely affects the adaptation performance (36.8\% and 37.3\% mAP) due to deterioration of domain gap in the graphical space. Besides, we find consistent performance improvements from 39.0\% to 43.9\% (3$^{rd}$ row to 6$^{th}$ row) with the increase of the node number from 20 to 200, because more nodes improve graph matching guided adaptation with better graphical space. However, using too many nodes (e.g., 500) will lead to the difficulty of graph matching optimization with a worse result (42.6\% mAP).
	
	\noindent\textbf{Evaluation on matching strategies.} We compare different settings between single-matching (each node is matched to the best counterpart) and multiple-matching (each node is matched to all counterparts in the same category) in Table~\ref{eq:match}. We find single-matching (43.5\% mAP$_{0.5}$) performs relatively better than multiple-matching (43.1\% mAP$_{0.5}$) because singe-matching aligns primary node pairs and relives noisy adaptation on ambiguous nodes. Besides, each component (TE, FS, QC) of the proposed matching loss contributes to the matching-based domain adaptation, yielding consistent mAP$_{0.5}$ improvements from 42.1\% to 43.5\%.
	
	\subsection{Qualitative Results}
	\noindent\textbf{Result comparison.} We present the comparison among (a) source only, (b) EPM~\cite{everypixelmatters}, (c) the proposed SIGMA and (d) ground-truth in Figure~\ref{show}. SIGMA can reduce missing errors, such as the truck in $1^{st}$ and $2^{nd}$ lines compared with the category-agnostic method EPM~\cite{everypixelmatters}. Besides, our approach also eliminates some classification errors (false-positive cases), such as the rider in $2^{nd}$ row, showing the advantage in category-level adaptation with well-aligned class-conditional distributions.
	
	\noindent\textbf{Feature comparison.} For each category, we randomly sample an equal number of pixels on ResNet-50-based features for each domain (200 pixels/ domain\&category) and present the T-SNE comparison with the GA baseline~\cite{everypixelmatters} in Figure~\ref{tsne}. It can be observed that those similar categories (\textit{person}, \textit{rider}, and \textit{bike}) can be separated clearly on features by our method, which benefits the followed detection head in terms of object recognition significantly.
	
	\section{Conclusion}
	In this paper, we propose a novel framework for DAOD, coined SIGMA. It represents domain information through semantic-complete graphs and model domain adaptation as a graph matching problem, which break the barrier of existing category-level approaches in terms of semantic mismatching and sub-optimal prototype alignment. It adopts a Graph-embedded Semantic Completion module (GSM) to complete mismatched semantics and model class-conditional distributions with graphs. Then, it leverages a Bipartite Graph Matching adaptor (BGM) to achieve fine-grained alignment with a node-to-node matching. Extensive experiments on three benchmarks show that the proposed method outperforms existing approaches significantly.
	
	\appendix \section{Sensitivity Analysis}
	
	\subsection{Parameter Sensitivity} As shown in Table~\ref{tab:para}, we analyze the sensitivity in terms of the adaptation intensity $\lambda_{1,2}$, where $\lambda_{1}$ works on the node classification loss and $\lambda_{2}$ controls the intensity of structure-aware matching loss. We first try a group of consistent parameters \{0.05, 0.1, 0.2\} for $\lambda_{1,2}$ (1$^{st}$ to 3$^{rd}$ lines), finding that decreasing the values leads to a significant performance drop compared with our main settings ($\lambda_{1,2}=0.1$). By fixing $\lambda_{1}$, increasing and decreasing $\lambda_{2}$ sightly decrease the overall performance, demonstrating that our setting ($\lambda_2=0.1$) is optimal. By fixing $\lambda_{2}$, decreasing $\lambda_{1}$ shows a significant negative impact on the framework while increasing it gives some further improvements. These results demonstrate that the larger intensity on the node loss contributes to establishing a better graphical space for the graph-matching-based adaptation.
	
	\begin{table}[h]
		\small
		\begin{center}
			\begin{tabular}{ c c |c c c}
				\toprule
				$\lambda_1$ & $\lambda_2$& mAP$_{0.5:0.95}$& mAP$_{0.5}$ &mAP$_{0.75}$\\
				\midrule
				0.05    &0.05               & 22.8 &42.2 &21.4   \\
				0.1     &0.1    &24.0&43.5&\textbf{23.5}\\
				0.2     &0.2                &\textbf{24.2}&43.3&23.3  \\
				\midrule
				0.1    &0.05                &23.2   &42.9   &23.0   \\
				0.1    &0.2               &23.5  &43.3 & 23.1 \\
				0.05    &0.1               &22.3   &42.0 & 21.8 \\
				0.2     &0.1                &\textbf{24.2}   &\textbf{43.7}& 23.1 \\
				\bottomrule
			\end{tabular}
		\end{center}
		\vspace{-7pt}
		\caption{\label{tab:para} Comparison results on Cityscapes$\to$Foggy Cityscapes (\%) of different settings of $\lambda_1$ and $\lambda_2$. We set $\lambda_{1,2}=0.1$ in the experiments of the manuscript as 2$^{nd}$ line. }
		\vspace{-10pt}
	\end{table}
	
	\subsection{Position Sensitivity} We further investigate the position to deploy the Node Discriminator (ND) to align the matched nodes, and record the comparison results in Table~\ref{tab: position}. We compare three settings for the node alignment, i.e., P1: semantic-complete nodes $\mathcal{V}_{s/t}$ (without the hallucination nodes), P2: enhanced nodes after graph convolution $\Tilde{\mathcal{V}}_{s/t}$, and P3: the nodes after Cross Graph Interaction (CGI) $\hat{\mathcal{V}}_{s/t}$. It can be observed that performing the alignment on the semantic-complete nodes (P1) achieves the best results with well-aligned node pairs. Besides, we find a significant performance drop on P3 because the proposed CGI will exchange information across domains, confusing the discriminator and harming the adversarial alignment. Hence, aligning nodes in P1 is optimal in the proposed method as the setting in our manuscript.

	\begin{table}[h]
		\small
		\begin{center}
			\begin{tabular}{p{0.45cm}<{\centering}  |p{0.45cm}<{\centering}  p{0.35cm}<{\centering}  p{0.35cm}<{\centering}  p{0.35cm}<{\centering}  p{0.35cm}<{\centering}  p{0.35cm}<{\centering}  p{0.35cm}<{\centering}
					p{0.5cm}<{\centering}|p{0.55cm}<{\centering} }
				\toprule
				Pos.& prsn &rider& car &truc &bus &train &moto &bike &mAP\\
				\midrule
				P1&\textbf{46.9}&\textbf{48.4}&\textbf{63.7}&27.1&50.7&35.9&\textbf{34.7}&\textbf{41.4}&\textbf{43.5}\\
				P2&43.9&46.0&57.0&\textbf{29.7}&\textbf{53.9}&\textbf{39.7}&34.6&39.6&43.0 \\
				P3&44.0&45.4&57.2&25.2&48.4&26.8&27.5&38.7&39.2\\
				\bottomrule
			\end{tabular}
		\end{center}
		\vspace{-10pt}
		\caption{Comparison results on Cityscapes$\to$Foggy Cityscapes (\%) by deploying the ND on different nodes, i.e., semantic-complete nodes $\mathcal{V}_{s/t}$ (P1), enhanced nodes after graph convolution $\Tilde{\mathcal{V}}_{s/t}$ (P2), and the nodes after cross graph interaction $\hat{\mathcal{V}}_{s/t}$ (P3).}
		\label{tab: position}
		\vspace{-10pt}
	\end{table}
	
	\subsection{Normalization Sensitivity~\label{norm}} The proposed method transforms the visual feature to the graphical space (V2G) with a projection module (Fc-Norm-ReLU-Fc). Hence, we present a comparison among different projection strategies with different normalization (Norm) tricks, including Group Normalization (commonly used in the FCOS~\cite{fcos} detection head), Batch Normalization (commonly used in the ResNet~\cite{resnet} backbone network), and Layer Normalization~\cite{LN}, as shown in Table~\ref{tab: norm}. Our projection design with Layer Normalization works better on node embedding than other common settings, preserving node-based correspondence and achieving the best adaptation result (43.5\% mAP).
	
	\begin{table}[h]
		\small
		\begin{center}
			\begin{tabular}{p{0.45cm}<{\centering}  |p{0.45cm}<{\centering}  p{0.35cm}<{\centering}  p{0.35cm}<{\centering}  p{0.35cm}<{\centering}  p{0.35cm}<{\centering}  p{0.35cm}<{\centering}  p{0.35cm}<{\centering}
					p{0.5cm}<{\centering}|p{0.55cm}<{\centering} }
				\toprule
				Pos.& prsn &rider& car &truc &bus &train &moto &bike &mAP\\
				\midrule
				GN&45.7&44.9&63.1&24.8&48.3&43.2&32.6&40.9&42.9\\
				BN&46.1&42.8&61.7&\textbf{27.6}&45.5&34.8&32.0&38.0&41.0 \\
				LN&\textbf{46.9}&\textbf{48.4}&\textbf{63.7}&{27.1}&\textbf{50.7}&\textbf{35.9}&\textbf{34.7}&\textbf{41.4}&\textbf{43.5}\\
				\bottomrule
			\end{tabular}
		\end{center}
		\vspace{-10pt}
		\caption{Comparison results on Cityscapes$\to$Foggy Cityscapes (\%) of different normalization strategies in the vision-to-graph (V2G) transformation.}
		\label{tab: norm}
		\vspace{-15pt}
	\end{table}
	
	\section{Discussion}
	\subsection{Baseline Selection}
	\noindent\textbf{Two-stage v.s. single-stage baselines.} Two-stage object detectors, e.g., Faster RCNN~\cite{fasterrcnn}, consist of a feature extractor, a Region Proposal Network (RPN) and a detection head for classification and regression. These approaches first adopt RPN on image features to obtain Region of Interests (RoIs), and then perform detection based on these region proposals. Differently, single-stage object detectors~\cite{fcos,yolov3} only contain a feature extractor and detection head, and these approaches directly make prediction on image features without RPN.
	
	\noindent{\textbf{Reasons for the singe-stage baseline.}}  In this paper, we mainly focus on the domain adaptation for singe-stage object detectors as lots of recently published works~\cite{SSAL,KTNet,everypixelmatters,DIDN,I3Net,simrod}, and we select the single-stage detector as the baseline because of the following two main reasons.
	
	\textbf{1) Discarding RPN.} Most adaptation works~\cite{xu2020crossgraph,RPN,c2f} perform adaptation on both image features and RoI representations, which highly rely on the RPN and are limited to the two-stage detectors. In contrast, our method achieves fine-grained adaptation only using image features and totally discards the RPN, yielding enormous potentials to be generalized to different baselines. Hence, we use the single-stage baseline free of RPN in our method to demonstrate the advantages without bells and whistles.
	
	\textbf{2) Fair comparison.} The fairness and agreement of the benchmark comparison have been proven in recently published literature~\cite{SSAL,KTNet,everypixelmatters,I3Net,simrod} for single-stage object detectors due to the comparable source only results and adaptation gains. Besides, we also report the fair adaptation gains in benchmark comparison to demonstrate our effectiveness in terms of domain adaptation. Moreover. most of the latest adaptation works~\cite{SSAL,KTNet,everypixelmatters,I3Net,simrod} are based on the single-stage detectors~\cite{fcos,yolov3}, and we aim to present a comparison with them using same baseline model.
	
	\noindent{\textbf{Potentials for the two-stage extension.}} We psropose a Graph-embedded Semantic Completion module (GSC) to complete the mismatched semantics and leverage a Bipartite Graph Matching adaptor (BGM) to achieve fine-grained adaptation on image features. These two modules are totally independent of the detection baseline types and can be effortlessly extended to different baselines by deploying on the features extracted from backbone networks.
	
	\subsection{Limitation}
\begin{figure}[h]
		\begin{center}
			\includegraphics[width=0.95\linewidth]{ 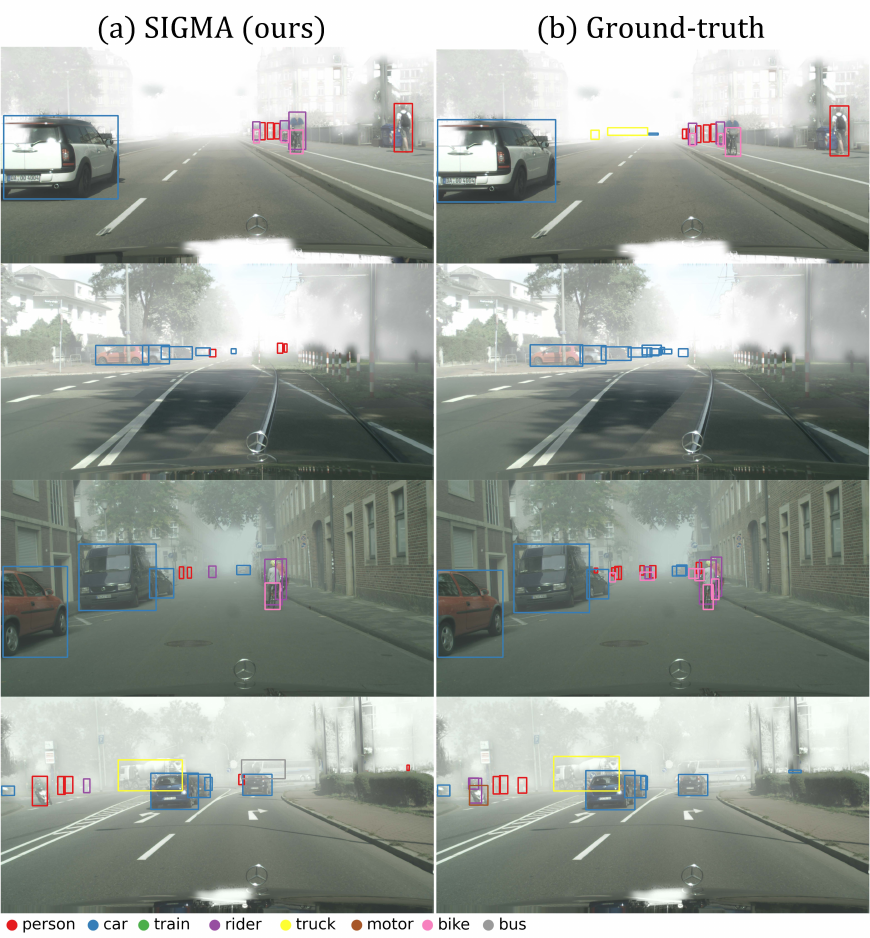}
		\end{center}
		\vspace{-10pt}
		\caption{Illustration of some failure examples compared between (a) the proposed SIGMA framework and (b) ground-truth. }
		\vspace{-10pt}
		\label{fail}
	\end{figure}

	Though the proposed model could achieve state-of-the-art results, it may have some failure cases (Figure~\ref{fail}) due to the limited visual features. As shown in 1$^{st}$ and 2$^{nd}$ row, we find that our method may miss and wrongly detect some distant objects obscured by heavy fog, e.g., the missing truck (1$^{st}$ row) and the wrongly detected person (2$^{nd}$ row), due to the poor visual features caused by the tiny scale (long distance) and low-quality appearance (heavy fog). This problem can be solved from two aspects, i.e., improving visual representations and compensating for visual features with other cues. On the one hand, we can use more robust backbone networks, e.g., ResNet-101~\cite{resnet}, to obtain better features than the VGG-16 backbone~\cite{vgg}. On the other hand, we can establish graph matching between visual and linguistic cues~\cite{ling} to compensate for the limited visual features with extra semantics.
	
	\section{Implementation Details}
	
	\subsection{Discriminator Architecture} As shown in Table~\ref{tab:more_arch_details}, we present the detailed architecture of the adversarial alignment module in our SIGMA framework, which includes the loss terms $\mathcal{L}_{GA}$ and $\mathcal{L}_{NA}$. We adopt image-level global alignment~\cite{DAfasterrcnn} using the Global Discriminator as~\cite{DAfasterrcnn,xu2020crossgraph,RPN,everypixelmatters,SSAL,DIDN,KTNet,Mega-DA}. Then, we introduce a node discriminator to align well-match graph nodes, as illustrated in the bottom part of Table~\ref{tab:more_arch_details}. Considering the graph nodes refactor the image-level spatial correspondence with edge connections, we replace the convolution layers with fully-connected layers. Besides, we change the Group Normalization (GroupNorm) with Layer Normalization (LayerNorm) due to the advantage of operating the node-based representation, as in Sec.~\ref{norm}.

	\begin{table}[h]
		\small
		\begin{center}
			\begin{tabular}{lll}
				\toprule
				\multicolumn{3}{c}{Global Discriminator~\cite{everypixelmatters}} \\
				\hline
				\multicolumn{3}{c}{Gradient Reversal Layer (GRL)} \\
				\multicolumn{3}{c}{Conv 256 $\times$ 3 $\times$ 3, stride 1 $\to$ GroupNorm $\to$ ReLU} \\
				\multicolumn{3}{c}{Conv 256 $\times$ 3 $\times$ 3, stride 1 $\to$ GroupNorm $\to$ ReLU} \\
				\multicolumn{3}{c}{Conv 256 $\times$ 3 $\times$ 3, stride 1 $\to$ GroupNorm $\to$ ReLU} \\
				\multicolumn{3}{c}{Conv 256 $\times$ 3 $\times$ 3, stride 1 $\to$ GroupNorm $\to$ ReLU} \\
				\multicolumn{3}{c}{Conv 1 $\times$ 3 $\times$ 3, stride 1}\\
				\midrule
				\multicolumn{3}{c}{Node Discriminator (ours)}   \\
				\hline
				\multicolumn{3}{c}{Gradient Reversal Layer (GRL)} \\
				\multicolumn{3}{c}{Fc 256 $\to$ LayerNorm $\to$ ReLU}\\
				\multicolumn{3}{c}{Fc 256 $\to$ LayerNorm $\to$ ReLU}\\
				\multicolumn{3}{c}{Fc 256 $\to$ LayerNorm $\to$ ReLU}\\
				\multicolumn{3}{c}{Fc 1   $\to$ LayerNorm $\to$ ReLU}\\
				\bottomrule
			\end{tabular}
		\end{center}
		\vspace{-10pt}
		\caption{\label{tab:more_arch_details}Architectures of the adversarial alignment modules.}
		\vspace{-15pt}
	\end{table}
	
	\renewcommand{\algorithmicrequire}{ \textbf{Input:}}
	\renewcommand{\algorithmicensure}{ \textbf{Output:}}
	\begin{algorithm}[t]
		\caption{Semantic-complete Graph Matching}
		\label{alg:SIGMA}
		\begin{algorithmic}[1]
			\REQUIRE~~\\
			$\mathcal{I}_{s/t}$: source and target images\\
			$\mathcal{Y}_{s}$: source annotations\\
			$\lambda_{1,2}$: hyperparameters in the loss function\\
			\ENSURE~~\\
			Domain adaptive object detector $\Theta$\\
			\FOR {$l=1$ {\bfseries to} $maxiter$}
			\STATE extract image features $\mathcal{F}_{s/t}$ witn backbone networks;\\
			\STATE generate global alignment loss $\mathcal{L}_{GA}$ on $\mathcal{F}_{s/t}$;\\
			\STATE send $\mathcal{F}_{s/t}$ to the detection head to generate $\mathcal{L}_{det}$ with $\mathcal{F}_{s}$ and classification maps $\mathcal{M}_{t}$ with $\mathcal{F}_{t}$;\\
			\emph{\textbf{Graph-embedded Semantic Completion (GSC)}}\
			\STATE perform V2G transformation to obtain nodes $\mathcal{V}^{raw}_{s/t}$; \\
			\STATE generate node alignment loss $\mathcal{L}_{NA}$; \\
			\STATE perform DNC for semantic-complete nodes $\mathcal{V}_{s/t}$; \\
			\STATE establish graphs $\mathcal{G}_{s/t}$ and perform GCN for $\Tilde{\mathcal{V}}_{s/t}$; \\
			\STATE update GMB with enhanced nodes $\Tilde{\mathcal{V}}_{s/t}$; \\
			
			\emph{\textbf{Bipartite Graph Matching (BGM)}}\
			\STATE perform CGI obtaining $\hat{\mathcal{V}}_{s/t}$ and generate loss $\mathcal{L}_{node}$;\\
			\STATE perform SNA matrix learning to obtain $\Tilde{\mathbf{M}}_{\mathrm{aff}}$;\\
			\STATE generate graph matching $\mathcal{L}_{mat}$;\\
			\emph{\textbf{Network Parameter Updating}}\
			\STATE use $\mathcal{L}= \lambda_{1}\mathcal{L}_{node} + \lambda_{2}\mathcal{L}_{mat} +\mathcal{L}_{NA}+ \mathcal{L}_{GA} + \mathcal{L}_{det}$ to update network parameters with backpropagation;
			
			\ENDFOR
			\RETURN Domain adaptive object detector $\Theta$;\\
		\end{algorithmic}
	\end{algorithm}
	
	\subsection{Implementation and Training}
	
	\textbf{1) Different blocks.} The non-linear projection layer used in the vision-to-graph (V2G) transformation is deployed with a Fc-LayerNorm-ReLU-Fc block, and the classifier for node classification is Fc-ReLU-Fc.
	
	\textbf{2) Dropout rate.} The dropout rate is set 0.1 for the edge-drop~\cite{dropedge} to avoid the potential visual bias.
	
	\textbf{3) Spectral clustering.} For the learning of the graph-guided memory bank, we perform spectral clustering if the number of nodes is larger than 5 to ensure the clustering reliability. Besides, we replace the Laplacian affinity~\cite{cluster} with K-Nearest Neighbor (K=5) in the clustering algorithm, which reduces the time-consuming significantly.
	
	\textbf{4) End-to-end training.} Our method can achieve end-to-end training without the warm-up stage. We utilize halved source nodes as the placeholders if no nodes appear in the target domain to train our matching module and introduce extra 10,000 iterations for training, which can achieve the same results as the warm-up-included strategy.
	
	\textbf{5) Multiple matching.} The detailed implementation of the multiple-matching ablation study (in Table 5 of our manuscript) is as follows,
	\begin{equation}
		\mathcal{L}_{mat} = Loss[sigmoid(\mathbf{M}_{\mathrm{aff}}), \mathbf{Y}_{\mathbf{\Pi}}],
	\end{equation}
	where $\mathbf{M}_{\mathrm{aff}}$ is the node affinity without adopting Instance Normalization and the Sinkhorn~\cite{Sinkhorn1964ARB} layer, and $Loss[A,B]$ can be selected as the BCE and MSE loss to evaluate the difference between $A$ and $B$.
	
	\subsection{Optimization Pipeline} The overall optimization pipeline of the proposed SIGMA framework is shown in Algorithem~\ref{alg:SIGMA}. Given the source and target images $\mathcal{I}_{s/t}$, source annotations $\mathcal{Y}_{s}$, and some predefined hyperparameters $\lambda_{1,2}$, we implement the SIGMA framework to obtain a domain adaptive object detector $\Theta$ with $maxiter$ iterative training.
	
	\begin{figure}[h]
		\begin{center}
			\includegraphics[width=0.9\linewidth]{ 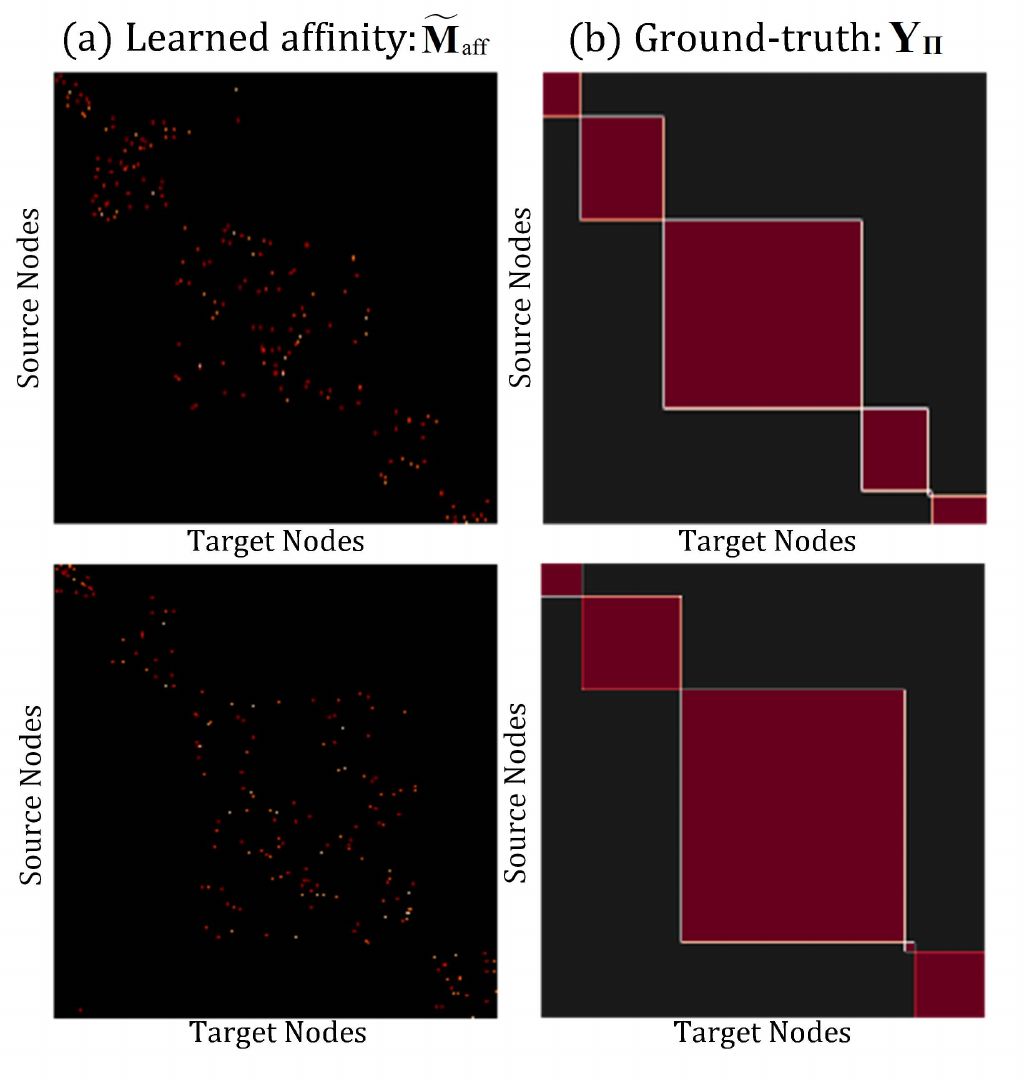}
		\end{center}
		\vspace{-18pt}
		\caption{Illustration of (a) the learned doubly stochastic affinity matrix $\Tilde{\mathbf{M}}_{\mathrm{aff}}$  and (b) the ground-truth $\mathbf{Y}_{\mathbf{\Pi}}$. Each activated entry $\Tilde{\mathbf{M}}_{\mathrm{aff}}^{i,j}$ represents an adaptive matching between the source node $\hat{v}^i_{s}$ and target node $\hat{v}^j_{t}$. Each positive entry $\mathbf{Y}_{\mathbf{\Pi}}^{i,j}$ (marked in red) indicates that the node $\hat{v}^i_{s}$ and $\hat{v}^j_{t}$ are in the same category.}
		\label{mat_show}
		\vspace{-10pt}
	\end{figure}
	
	\section{Qualitative Results}
	\subsection{Matching Visualization}
	As shown in Figure~\ref{mat_show}, we visualize the learned doubly stochastic node affinity matrix $\Tilde{\mathbf{M}}_{\mathrm{aff}}$ and the ground-truth matrix $\mathbf{Y}_{\mathbf{\Pi}}$ (Refer to Figure~2 of the manuscript for better understanding.). Each activated entry $\Tilde{\mathbf{M}}^{i,j}_{\mathrm{aff}}$ represents a matched node pair across domains, and each activated entry $\mathbf{Y}^{i,j}_{\mathbf{\Pi}}=1$ (marked in red) indicates that the source node $\hat{v}^i_s$ and the target counterpart $\hat{v}^j_t$ are in the same category. Based on the proposed structure-aware matching loss, each source node successfully find an optimal target node in the same category adaptively and match it to achieve graph-matching-based adaptation.
	
	\subsection{Qualitative Comparison}
\begin{figure*}[h]
		\begin{center}
			\includegraphics[width=0.95\linewidth]{ 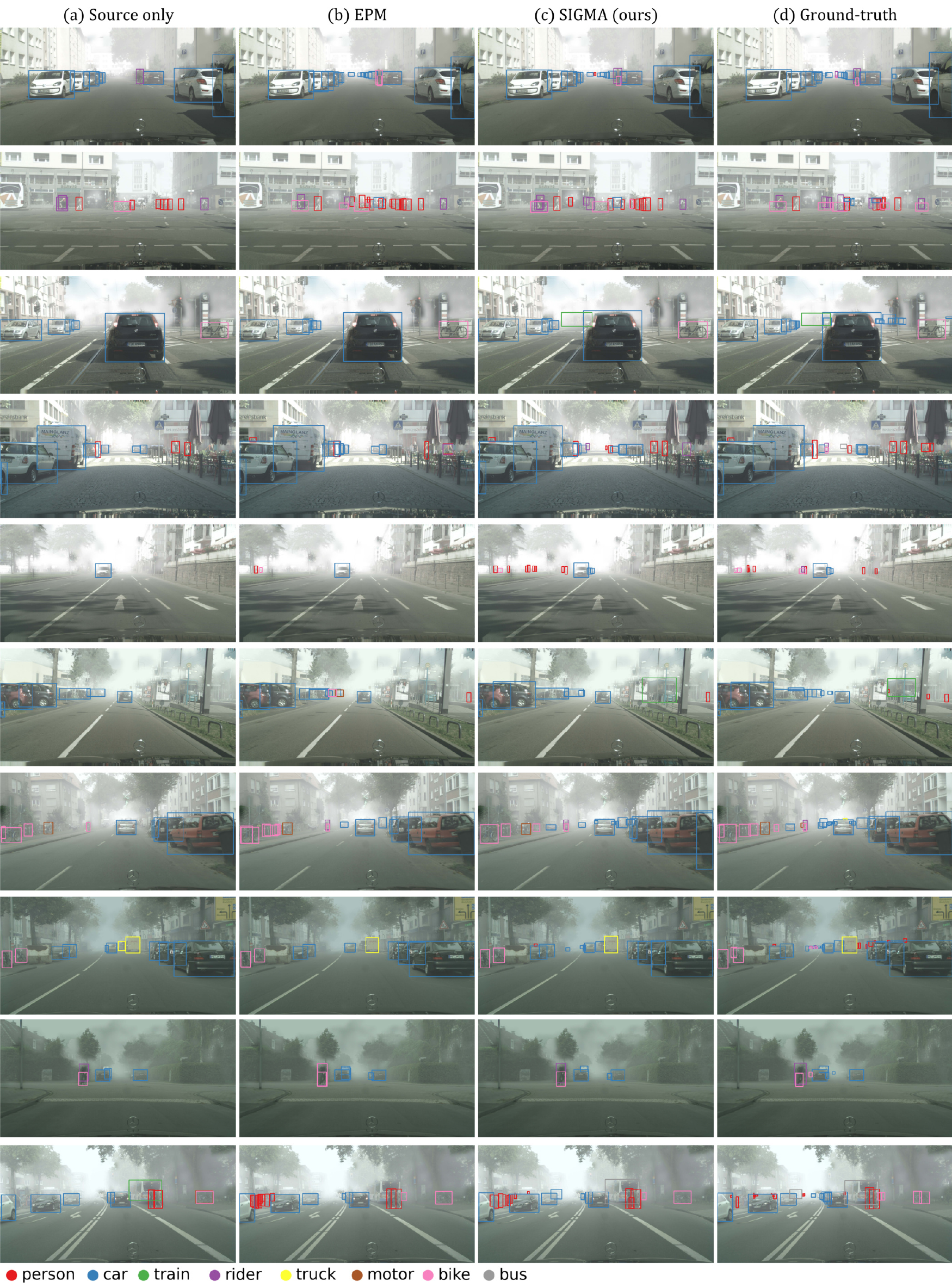}
		\end{center}
		\vspace{-18pt}
		\caption{Qualitative reustks on the {Cityscapes$\to$Foggy Cityscapes} adaptation scenario of (a) the source only model, (b) EPM~\cite{everypixelmatters}, (c) the proposed SIGMA, and (d) Ground-truth. (Zooming in for best view.) }
		\label{show1}
	\end{figure*}

	We present more qualitative comparisons among (a) source only, (b) EPM~\cite{everypixelmatters}, (c) the proposed SIGMA, and (d) ground-truth in Figure~\ref{show1}. Our method can eliminate some missing errors (false-negative cases) and avoid some wrong classification cases (false-positive cases) compared with the class-agnostic method EPM~\cite{everypixelmatters}, which verifies the effectiveness of aligning class-conditional distributions.
	
	\clearpage
	{\small
		\bibliographystyle{ieee_fullname}
		\bibliography{egbib}
	}
	
\end{document}